\documentclass[twoside,twocolumn]{article}

\usepackage{blindtext} 

\usepackage[sc]{mathpazo} 
\usepackage[T1]{fontenc} 
\usepackage{microtype} 

\usepackage[english]{babel} 

\usepackage[hmarginratio=1:1,top=32mm,columnsep=20pt]{geometry} 
\usepackage[hang, small,labelfont=bf,up,textfont=it,up]{caption} 
\usepackage{booktabs} 

\usepackage{lettrine} 

\usepackage{enumitem} 
\usepackage{graphicx}
\DeclareGraphicsExtensions{.pdf,.png,.jpg}
\usepackage{enotez}
\setenotez{list-name={References}}
\usepackage{hyperref}

\setlist[itemize]{noitemsep} 

\usepackage{abstract} 

\usepackage{titlesec} 
\renewcommand\thesection{\Roman{section}} 
\renewcommand\thesubsection{\roman{subsection}} 
\titleformat{\section}[block]{\large\scshape\centering}{\thesection.}{1em}{} 
\titleformat{\subsection}[block]{\large}{\thesubsection.}{1em}{} 

\usepackage{fancyhdr} 
\usepackage{titling} 

\usepackage{hyperref} 

\makeatletter
\newcommand{\printfnsymbol}[1]{%
  \textsuperscript{\@fnsymbol{#1}}%
}
\makeatother
\pretitle{\begin{center}\Huge\bfseries} 
\posttitle{\end{center}} 
\title{A Study on Deep Learning Based Sauvegrain Method for Measurement of Puberty Bone Age} 
\author{%
\textsc{Seung Bin Baik}\thanks{These two authors contributed equally} \\[1ex] 
\normalsize Plani Inc. \\ 
\normalsize \href{mailto:bsb@plani.co.kr}{bsb@plani.co.kr} 
\and 
\textsc{Keum Gang Cha}\printfnsymbol{1} \\[1ex] 
\normalsize Plani Inc. \\ 
\normalsize \href{mailto:chagmgang@plani.co.kr}{chagmgang@plani.co.kr} 
}

\date{\today} 
\renewcommand{\maketitlehookd}{%
\begin{abstract}
\noindent  
This study applies a technique to expand the number of images to a level that allows deep learning. And the applicability of the Sauvegrain method through deep learning with relatively few elbow X-rays is studied.  The study was composed of processes similar to the physicians' bone age assessment procedures. The selected reference images were learned without being included in the evaluation data, and at the same time, the data was extended to accommodate the number of cases. In addition, we adjusted the X-ray images to better images using U-Net and selected the ROI with RPN + so as to be able to perform bone age estimation through CNN. The mean absolute error of the Sauvegrain method based on deep learning is 2.8 months and the Mean Absolute Percentage Error (MAPE) is 0.018. This result shows that X - ray analysis using the Sauvegrain method shows higher accuracy than that of the age group of puberty even in the deep learning base. This means that deep learning of the Suvegrain method can be measured at a level similar to that of an expert, based on the extended X-ray image with the image data extension technique. Finally, we applied the Sauvegrain method to deep learning for accurate measurement of bone age at puberty. As a result, the present study is based on deep learning, and compared with the evaluation results of experts, it is possible to overcome limitations of the method of measuring bone age based on machine learning which was in TW3 or Greulich \& Pyle due to lack of X- I confirmed the fact. And we also presented the Sauvegrain method, which is applicable to adolescents as well.
\end{abstract}
}

\begin{document}

\maketitle


\section{Introduction}

\lettrine[nindent=0em,lines=3]{A}s entered modern society, the interactions of different cultures with one another increased. In particular, improvements and innovations in logistics and distribution have made it possible to exchange food, including food. In the case of Oriental, the living environment changed drastically due to the influx of western eating habits, which causes new problems such as gastrophy of children and adolescents. The importance of measurement of bone age has emerged in order to prevent this and appropriate treatment, and the demand for evaluation of growth and development of children and adolescents is increasing. This bone age measurement is used to assess the extent of growth and development.\endnote{Vallejo-Bolaños E, España-López AJ, Muñoz-Hoyos A, Fernandez-Garcia JM. The relationship between bone age, chronological age and dental age in children with isolated growth hormone deficiency. Internaltional Journal of Paediatric Dentistry. 1999: 9(3): 201-206} Most doctors are still evaluating age-standard and x-ray images one by one using their eyes. Due to these external factors, doctors are spending more time measuring bone age, and there is an increasing need to improve working time, which is about 5 minutes per work.
\\*\\*
In recent years, studies have been actively conducted to utilize image processing capabilities of computers in order to solve these problems. In particular, Computer-assisted bone age assessment: Image preprocessing and epiphyseal/metaphyseal ROI extraction\endnote{Pietka, Ewa, et al. "Computer-assisted bone age assessment: Image preprocessing and epiphyseal/metaphyseal ROI extraction." IEEE transactions on medical imaging 20.8 (2001): 715-729} and Method and program for bone age calculation using deep neural networks\endnote{Asan Foundation and VUNO, Method and program for bone age calculation using deep neural networks. WO2017022908A1, filed March 2, 2016, issued April 2, 2018} are represetative. However, previous studies, including these studies, used published data from contests such as the 2017 RSNA Bone Age Challenge. Therefore, it was limited to the use of wrist x-ray images, and it was researched and developed focusing on the TW3 method. In the previous study, the Greulich \& Pyle and TW3 methods\endnote{Lloyd L Morris. Assessment of Skeletal Maturity and Prediction of Adult Height(TW3 Method). 2003: 47(3): 340-341} used to measure bone age have the disadvantage that it is difficult to read accurately. This is because the growth rate is increased during the two-year period when the sex-hormone and growth hormone secretion increase from 7.5 ~ 5.5cm/year to 7.5 ~ 8.5cm/year in girls aged between 11-13 years and in boys aged 13-15 years.\endnote{Diméglio A, Charles YP, Daures JP, de Rosa V, Kaboré B. Accuracy of Sauvegrain method in determining skeletal age during puberty. Journal of Bone and Joint Surgery. 2005: 87(8):1689-96} Dimeglio A suggests that the use of Sauvegrain method at puberty is highly accurate and actively utilized.\endnote{Luis Perez, Jason Wang. The Effectiveness of Data Augmentation in Image Classification using Deep Learning. arXiv 2017 Dec 13} However, since Sauvegrain method using elbow X-ray has not been fully disclosed in X-ray image, computer can be learned The amount of data is not sufficient. This can not accurately measure the bone age of adolescents, and previous studies have solved some of the problems that they are trying to solve, or suggested only the possibility. They did not provide a meaningful measure of bone age for all ages, including puberty.
\\*\\*
The purpose of this study is to investigate the applicability of the Sauvegrain method through in-depth learning using only the elbow x-ray images by applying The Effectiveness of Image Augmentation in Image Classification using Deep Learning\endnote{Kyung Mo, Yung. The Study on Standard Bone Age and Normal Range in Korean Children. Journal of the Korean Radiological Society. 1996: 34(2) : 269-276}. It is also expected that this will solve the remaining problems of existing problems in bone age measurement by first introducing a bone age measurement method based on deep learning at puberty.


\section{Methods}

This study is based on the general procedure of physician performing bone age measurement. The flow of the whole study is shown in Fig. At this time, in the case of the deep learning model, it is constituted sequentially according to the process that is commonly used. 

\begin{figure}[htbp]
\begin{center}
    \includegraphics[scale=0.25]{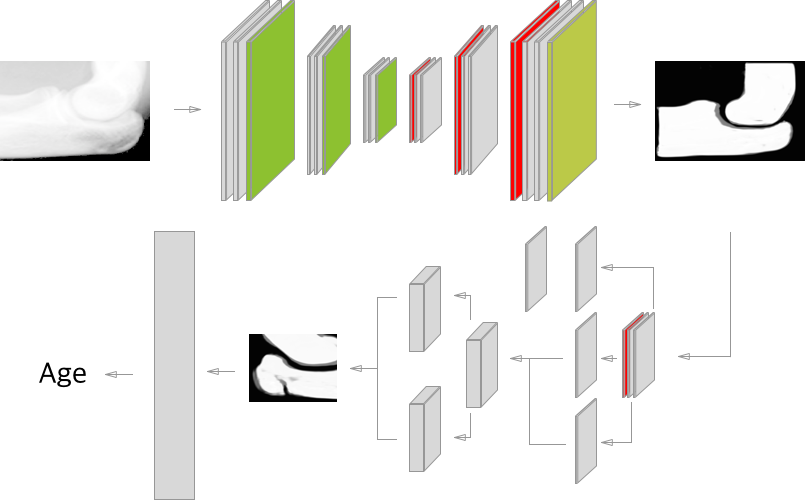}
    \caption{Overall Deep Learning Architecture}
\end{center}
\end{figure}
To compare the reference X-ray images, we selected the age of bone presented in the 10 elbow x-ray images of A COMPARISON OF THE SIMPLIFIED OLECRANON AND DIGITAL METHODS OF ASSESSMENT OF SKELTAL MATURITY DURING THE PUBERTAL GROWTH SPURT\endnote{Canavese, F., et al. "A comparison of the simplified olecranon and digital methods of assessment of skeletal maturity during the pubertal growth spurt." The bone \& joint journal 96.11 (2014): 1556-1560}. At this time, the original images were excluded from the learning images to be used as evaluation images at the end of the research. Also, elbow x-ray images of 10 and 11 year old boys missing from ' A comparison of the simplified olecranon and digital methods of assessment of skeltal maturity during the pubertal growth spurt' were added as reference images of elbows of 10 and 11 year old boys presented in STUDY OF SECONDARY OSSIFICATION CENTERS OF THE ELBOW IN THE BRAZILIAN POPULATION\endnote{Miyazaki, Cesar Satoshi, et al. "Study of secondary ossification centers of the elbow in the brazilian population." Acta ortopedica brasileira 25.6 (2017): 279-282}. After this procedure, the reference X-ray image was finally defined as shown in Fig 2.
\begin{figure}[htbp]
\begin{center}
    \includegraphics[scale=0.5]{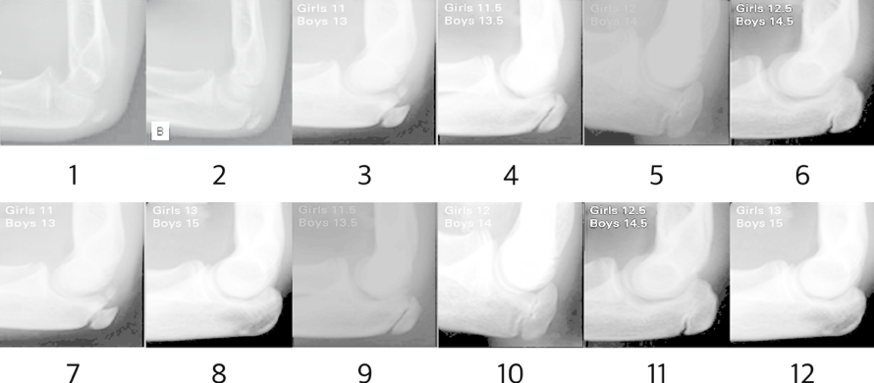}
    \caption{Test Data for Elbow Image}
\end{center}
\end{figure}
Since the researchers lacked the number of X-ray images, it was difficult to cope with the various cases where the actual data were input, so the reference images were enlarged using clipping, inversion, and rotation. The process of extension was performed as follows. Moving and cutting vertically and horizontally at every 10 pixels were performed and rotated 15 degrees and left and right reversed images were used as images for inputting a total of 576 images into the neural network. As a result, the study was conducted as shown in Fig. 3, and a image based on the standard age was used as the learning data, and the image with the label and the image without the learning were used as the validation data.
\begin{figure}[htbp]
\begin{center}
    \includegraphics[scale=0.5]{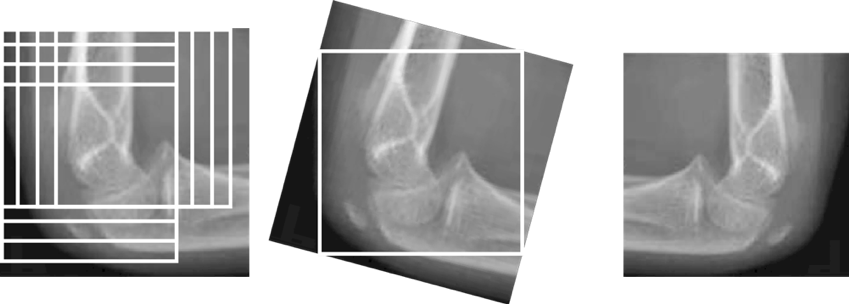}
    \caption{Image Augmentation}
\end{center}
\end{figure}
In order to increase the accuracy of the analysis, all of the above-mentioned data were subjected to removal of the skin region represented in the X-ray image and to more clearly distinguish the region corresponding to the bone. This is due to the fact that there is a risk that the necessary information may be lost due to unnecessary information in the X-ray image. Therefore we used, U-Net: Convolutional Networks for Biomedical Image Segmentation\endnote{Olaf Ronneberger, Philipp Fischer, and Thomas Brox. U-Net: Convolutional Networks for Biomedical Image Segmentation. arXiv. 18 May 2015}, which is a widely used Variable Auto Encoder(VAE) neural network, is used for medical image segmentation. U-Net is known to be well suited for medical image processing because it can distinguish between skin and bone image areas. It is known in a simpler and more efficient way with an end-to-end structure. Like other VAE neural networks, U-Net consists of a nested integral encoder and decoder called the contracting path and an expanding path. When up-sampling in the extension path, it works by finding the basic information by connecting the features of the contraction path.
\begin{figure}[htbp]
\begin{center}
    \includegraphics[scale=0.4]{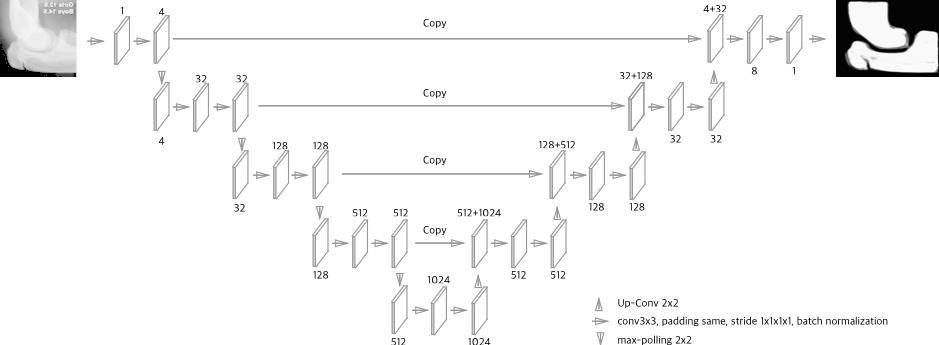}
    \caption{Unet Architecture}
\end{center}
\end{figure}
In general, the reference images were selected according to the process of bone age measurement, and well-separated X-ray images output from the above-described U-Net results were used as reference images. Finally, the next step was to unify the size of the image recommended by U-Net to 720x480 in grayscale to maintain image connectivity.
After extracting the bone region from the X-ray image, the RPN was performed in the flow as shown in Fig. 5, and the focused image was obtained in the desired region. At this time, the joint of the elbow is most important for the measurement of the bone age, and the bone age can be more efficiently measured by concentrating on the desired site. Therefore, we used RPN+(Region Proposal Network Plus), a modified version of Fast Regional-based Convolutional Network (F-RCNN)\endnote{Shiyu Huang, Deva Ramanan. Expecting the Unexpected: Training Detectors for Unusual Pedestrians with Adversarial Imposters. The IEEE Conference on Computer Vision and Pattern Recognition. 2017}\endnote{Ross Girshick. Fast R-CNN. Nerual Information Processing Systems. 2015} to extract the area of interest we intend to analyze. At this time, CNN (Convolutional Neural Network) was used to extract the area of interest.\endnote{Zakariya Qawaqneh(1), Arafat Abu Mallouh(1), Buket D. Barkana(2) Deep Convolutional Neural Network for Age Estimation based on VGG-Face Model. arXiv 2017 Sep 6}
\begin{figure}[htbp]
\begin{center}
    \includegraphics[scale=0.5]{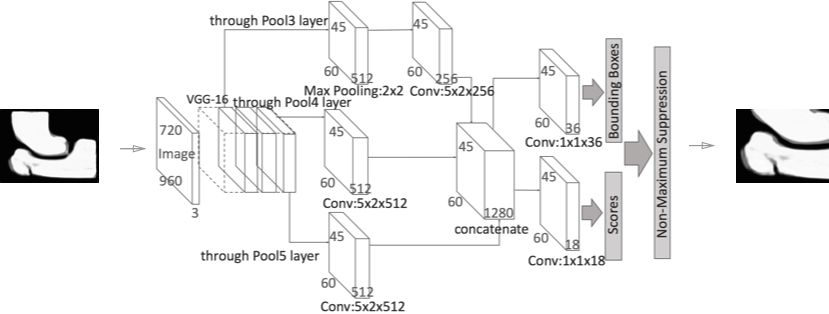}
    \caption{RPN+ Architecture}
\end{center}
\end{figure}
In this study, we use a generative hostile neural network to generate ROI (Region of Interest) in images. In this neural network, the VGC16 neural network is trained to output the probability that the input image will be true.\endnote{Seokhwan Kim, Rafael E. Banchs, Haizhou Li. Exploring Convolutional and Recurrent Neural Networks in Sequential Labelling for Dialogue Topic Tracking. Proceedings of the 54th Annual Meeting of the Association for Computational Linguistics. 2016} Local Proposals We have learned to return ROI boxes directly rather than learning RPNs to return objective proposals using a neural network based detection system. This is also called RPN +. In order to input imgaes into RPN +, I adjusted the size of images by rotating 90 degrees so as to have the same 1.5 magnification to adjust the size of the images to be input from the previously unified 720x480 to 720x960. Based on this, the representative age group is classified into one class based on the representative images of the representative age by age, and the similarity between the images and the representative images is measured and the corresponding age range is estimated through the CNN regression analysis as shown in Fig 6.
\begin{figure}[htbp]
\begin{center}
    \includegraphics[scale=0.5]{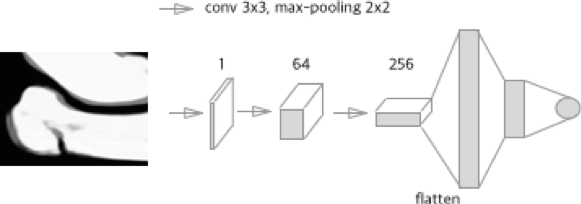}
    \caption{Convolutional Neural Network Architecture}
\end{center}
\end{figure}


\section{Results}
For the evaluation of the results of this study, the first set of original image data not used in actual learning was used as the evaluation protocol. Table 1 shows the data used, the age measured by the experts, and the age estimated by the system. The ages measured by the experts are the bone age measured by the experts on the study with each image.
\begin{table}[h]
\centering
\begin{tabular*}{1\linewidth}{@{\extracolsep{\fill}}ccc}
X-ray Number & Expert & System \\ \hline
1 & 120 & 126 \\ \hline
2 & 132 & 132 \\ \hline
3 & 156 & 151.2 \\ \hline
4 & 162 & 168 \\ \hline
5 & 168 & 168 \\ \hline
6 & 174 & 174 \\ \hline
7 & 156 & 162 \\ \hline
8 & 180 & 175.2 \\ \hline
9 & 162 & 159.6 \\ \hline
10 & 168 & 169.2 \\ \hline
11 & 174 & 176.4 \\ \hline
12 & 180 & 180 \\ \hline
\end{tabular*}
\caption{Comparison, the age estimated by expert and system in each method}
\end{table}
\\*\\*
The mean absolute error of the Sauvegrain method based on in-depth learning is 2.8 months and the Mean Absolute Percentage Error (MAPE) is 0.018. This result shows that the Sauvegrain method is applied enough for deep learning even though the image data extension technique is applied. This means that deep learning of the Suvegrain method can be measured at a level similar to that of an expert, based on the extended X-ray image with the image data extension technique.

\section{Discussion}
In this study, we describe the difficulty of measuring bone age at puberty, and the application of deep learning in deficient data. As a result, the Sauvegrain method was applied to accurately measure bone age at puberty. At this time, it was necessary to solve the problem of lack of data which can cope with the number of various cases, and data extension technique was applied. We measured bone age based on deep learning and compared it with the results of the experts. As a result, the Sauvegrain method, which can be applied to the puberty age beyond the limit of the machine learning based bone age measurement method, which stayed in TW3 or Greulich \& Pyle due to the lack of X-ray image, was presented for the first time.
\\*\\*
And there was no X-ray of the elbows and wrists of the same person. However, if we assume that an object can utilize the ensemble method composed of TW3 and Sauvegrain method, we expect to be able to measure bone age in a wider age group.

\clearpage
\printendnotes


\end{document}